\documentclass{article}

\usepackage[final]{corl_2026} 

\usepackage{graphicx}
\usepackage{amsmath}
\usepackage{amssymb}
\usepackage{amsfonts}
\usepackage{amsthm}
\usepackage{mathtools}
\usepackage{natbib}
\usepackage{multirow}
\usepackage{xcolor}
\usepackage{balance}
\usepackage{bbm}
\usepackage{booktabs}
\usepackage{algorithm}
\usepackage{algpseudocode}
\usepackage{enumitem}
\usepackage{microtype}

\newtheorem{problem}{Problem}

\newtheorem{proposition}{Proposition}

\title{Adaptive Risk-Certified Event-Triggered Replanning for Dynamic Navigation}

\author{
  Richie R. Suganda\\
  Department of Electrical Engineering and Computer\\
  University of Houston\\
  Houston, TX 77004, USA\\
  \texttt{rrsugand@cougarnet.uh.edu} \\
  \And
  Bin Hu \\
  Department of Engineering Technology \\
  University of Houston\\
  Houston, TX 77004, USA\\
  \texttt{bhu11@central.uh.edu} \\
}

\begin{document}
\maketitle

\begin{abstract}
Safe navigation in dynamic environments requires robots to plan under obstacle predictions whose errors are uncertain, non-stationary, and can induce rare but safety-critical failures. Existing control-barrier-function safety filters can reject immediately unsafe controls, but they provide little guidance on when the current finite-horizon planning mode itself is becoming unsafe as prediction uncertainty evolves. We propose Conformal Event-Triggered Risk-Certified Replanning (\emph{CERT-Replan}), a framework that uses calibrated barrier risk as an early-warning signal for replanning. CERT-Replan calibrates horizon-indexed obstacle-prediction residuals online and uses the resulting uncertainty radii to evaluate dynamic-obstacle safety margins. A one-step safety filter protects the next applied control, while a horizon-level risk monitor evaluates the upper-tail CVaR of predicted barrier-violation losses along the current MPC rollout. When this risk exceeds an allocated budget, CERT-Replan rejects the current planning mode and selects a lower-risk alternative, such as a different speed profile, corridor, or homotopy class, rather than repeatedly correcting the same nominal plan. In a non-stationary benchmark, CERT-Replan achieves an \(83.3\%\) collision reduction relative to the safety-filter-only baseline \textcolor{black}{and a \(77.8\%\) reduction relative to simple replanning triggers}, while reducing average safety-filter intervention by \(32.4\%\). \textcolor{black}{With Trajectron++, CERT-Replan achieves \(96\%\) collision-free operation. Hardware experiments and onboard runtime profiling demonstrate computational feasibility.}
\end{abstract}


\section{Introduction}
\label{sec:introduction}

Autonomous robots operating in crowds, warehouses, sidewalks, and mixed-traffic environments must make local planning decisions from imperfect predictions of nearby agents. A typical navigation stack estimates obstacle states from onboard sensing, predicts short-horizon obstacle motion, computes a local plan, and applies a safety filter before execution. Certifying this pipeline is challenging because obstacle-prediction errors are rarely stationary or uniform: they can grow under occlusion, close interaction, dense traffic, abrupt intent changes, and changes in sensing geometry. Modern multi-agent predictors capture rich interaction patterns, but their uncertainty can be multi-modal and can shift as the robot enters new sensing or interaction regimes~\cite{alahi2016social,gupta2018social,salzmann2020trajectron++,ngiam2021scene,shi2022motion,zhou2023query,cheng2023forecast,tumu2024multi}. As a result, a nominal forecast may be accurate in benign scenes but unreliable in rare, safety-critical tail events.

Control barrier functions (CBFs) provide a principled optimization-based mechanism for rejecting immediately unsafe controls~\cite{ames2019control}. For dynamic-obstacle navigation, however, the safety margin depends not only on the robot state but also on uncertain future obstacle states. Dynamic parabolic CBFs (DPCBFs) improve dynamic-obstacle avoidance by incorporating relative velocity in a line-of-sight frame, thereby reducing the conservatism of purely distance-based or collision-cone-like constraints in dense scenes~\cite{park2025beyond}. Robust, distributionally robust, and perception-aware CBFs further address model, sensing, and obstacle-state uncertainty~\cite{dacs2025robust,long2026sensor,zhang2024online,suganda2025distributed}. Yet these filters are fundamentally local in time: they protect the next executed input, but a nominal or one-step filtered control does not by itself indicate whether the current finite-horizon planning mode is becoming unsafe as prediction uncertainty evolves.

Risk-aware safety filters partially address this issue by replacing pointwise safety constraints with constraints on the distribution of safety margins. Conditional value-at-risk (CVaR) barrier functions protect against low-probability, high-severity events by constraining the adverse tail of the safety-margin distribution~\cite{ahmadi2021risk,akella2025risk}, and risk-adaptive CVaR methods tune the risk level online to balance conservativeness and feasibility~\cite{wang2025safe}. These methods, however, still require uncertainty samples or error models that remain representative under distribution shift. Adaptive conformal inference (ACI) provides a complementary mechanism by recalibrating uncertainty online from recent prediction residuals without assuming a fixed parametric error model~\cite{gibbs2021adaptive,dixit2023adaptive}. Conformal planning has been used to construct calibrated prediction regions for dynamic environments~\cite{lindemann2023safe}, and recent work has extended conformal ideas to adaptive POMDP planning, recursively feasible MPC, and time-aware planning in dynamic scenes~\cite{sheng2024safe,stamouli2024recursively}. In this paper, we use online conformal calibration as the data-driven adaptation mechanism that supplies horizon-indexed uncertainty radii to both immediate safety filtering and horizon-level plan evaluation.

The key gap we address is temporal. A one-step safety filter can repeatedly modify unsafe controls, but repeated correction of the same nominal plan may still push the robot toward regions where future feasible controls are scarce or unavailable. Similarly, resolving the same MPC problem at every step does not answer the higher-level question of when the robot should change the local reference, speed profile, corridor, homotopy class, or risk allocation. Event-triggered replanning provides the appropriate abstraction: update the plan only when a task-relevant condition indicates that the current mode should no longer be trusted~\cite{tabuada2007event,eqtami2013event,honda2024when}. Our central idea is to use calibrated barrier risk as this condition. Rather than treating risk only as a constraint on the next input, CERT-Replan uses it as an early-warning signal for replacing the current local planning mode.

This paper makes four contributions:
\begin{enumerate}
\item We introduce a horizon-indexed adaptive conformal calibration mechanism for dynamic-obstacle prediction errors, producing online position and velocity uncertainty radii that are shared by immediate safety filtering and horizon-level mode evaluation.
\item We formulate an ACI+Adaptive-CVaR-DPCBF safety filter that uses calibrated geometric tightening and lower-tail CVaR constraints to protect the next applied control under non-stationary prediction uncertainty, while adapting the CVaR tail mass to preserve feasibility.
\item We propose a finite-horizon CERT risk monitor that evaluates the current local planning mode through upper-tail CVaR of predicted barrier-violation losses, yielding a conditional risk certificate for the evaluated mode.
\item We integrate this certificate into an event-triggered MPC replanning layer that replaces the current mode when its allocated horizon-risk budget is exceeded, and validate the resulting framework in randomized simulations and hardware-in-the-loop experiments.
\end{enumerate}

\section{Related Work}
\label{sec:related}

\paragraph{Local planning and replanning.}
Classical reactive and local-planning methods include dynamic-window approaches, graph replanning, sampling-based motion planning, and trajectory optimization~\cite{fox1997dynamic,stentz1995focussed,karaman2011sampling,rosmann2017integrated,ratliff2009chomp,kalakrishnan2011stomp}. These methods can react to dynamic scenes, but they typically require additional uncertainty handling to provide probabilistic safety under prediction errors. Risk-aware planning methods incorporate chance constraints, distributional robustness, coherent risk measures, or sampling-based risk estimates to account for uncertain obstacle motion~\cite{renganathan2023risk,dixit2023risk,trevisan2025dynamic}. Event-triggered and self-triggered control methods update communication or control actions only when needed to preserve system-level properties under resource or uncertainty constraints~\cite{tabuada2007event,hu2014event,hu2021stochastic}. Event-triggered MPC and learned replanning policies reduce unnecessary replans and can improve navigation efficiency, but existing triggers are often based on tracking error, feasibility, map updates, or learned heuristics rather than calibrated tail risk~\cite{eqtami2013event,honda2024when}. In contrast, our trigger is tied to a calibrated finite-horizon risk certificate for the current local planning mode.

\paragraph{Prediction uncertainty and conformal planning.}
Multi-agent trajectory prediction is often multi-modal due to latent goals and interactions~\cite{alahi2016social,gupta2018social,salzmann2020trajectron++,ngiam2021scene,shi2022motion,zhou2023query,cheng2023forecast}. Conformal prediction gives distribution-free finite-sample coverage under exchangeability~\cite{shafer2008tutorial,angelopoulos2023conformal}, while adaptive conformal inference updates the miscoverage level online to handle distribution shift~\cite{gibbs2021adaptive}. In robot motion planning, conformal prediction has been used to construct prediction regions for dynamic obstacles~\cite{lindemann2023safe}, adaptive conformal prediction has been used for online planning among dynamic agents~\cite{dixit2023adaptive,suganda2026formation}, and recent work has extended conformal ideas to time-aware dynamic-environment planning and CBF safety certification~\cite{liang2025time,zhang2025conformal}. Our method differs by using calibrated residuals inside a risk-sensitive barrier filter and by using horizon-level tail risk as a replanning trigger.

\paragraph{Risk-aware CBFs.}

CBFs have become a standard tool for safety-critical control because they can be embedded in optimization-based controllers~\cite{ames2019control,garg2024advances}. CVaR-CBFs extend this idea to stochastic safety and protect against poor-tail outcomes~\cite{ahmadi2021risk}. Risk-adaptive CVaR barrier functions adjust the CVaR level online to avoid unnecessary conservativeness while preserving feasibility~\cite{wang2025safe}. Recent safety-filter and predictive-control methods combine CBFs with differentiable optimization, sampling-based MPC, learned neural barriers, and conformal prediction for uncertainty-aware safety certification~\cite{DaiKKGTK23,yin2023shield,so2024train, zhang2025conformal}. DPCBFs improve dynamic-obstacle avoidance by shaping the safe set with both relative position and relative velocity~\cite{park2025beyond}. Closely related MPC-CBF and predictive-CBF methods encode barrier conditions over a prediction horizon and provide feasibility or safety analyses under modeled dynamics and constraints~\cite{zeng2021safety,wabersich2023predictive,breeden2022predictive}. Our framework is complementary: rather than proposing a new nominal DPCBF or MPC-CBF formulation, we use adaptive conformal calibration and CVaR barrier risk to decide when the current local planning mode should be replaced under non-stationary obstacle-prediction uncertainty.

\section{Problem Formulation}
\label{sec:problem}

This section formalizes the dynamic navigation problem considered in this work.
A robot must navigate among multiple moving obstacles using finite-horizon
obstacle forecasts whose error distributions are unknown, non-stationary, and
not assumed to follow a fixed parametric model. The objective is to design an
online navigation policy that maintains task progress while enforcing
risk-aware safety and replanning before the current local plan becomes unsafe.

\noindent\textbf{System and obstacle prediction.}
We consider a robot with discrete-time dynamics
\begin{equation}
    x_{k+1}=F(x_k,u_k),
    \qquad
    x_k\in\mathcal X,\quad u_k\in\mathcal U,
    \label{eq:robot_dynamics}
\end{equation}
where $x_k$ is the robot state, $u_k$ is the control input, and $\mathcal U$ is
the admissible input set. At time $k$, the robot observes a set of dynamic
obstacles indexed by $\mathcal I_k$. Obstacle $j\in\mathcal I_k$ has measured
state $o_k^j=[p_{j,k}^{o\top},v_{j,k}^{o\top}]^\top\in\mathcal O$, where
$p_{j,k}^o,v_{j,k}^o\in\mathbb R^2$ denote its measured position and velocity. The robot is provided with a finite-horizon obstacle predictor
\begin{equation}
    \widehat{\mathbf O}_k^j
    =
    \Pi_\theta(o_k^j)
    =
    \left(\hat o^j_{k+1|k},\ldots,\hat o^j_{k+N|k}\right),
    \qquad
    \hat o^j_{k+\tau|k}
    =
    [\hat p_{j,k+\tau|k}^{o\top},\hat v_{j,k+\tau|k}^{o\top}]^\top .
    \label{eq:predictor}
\end{equation}
The predictor may use additional information such as obstacle history, robot
state, map context, or interaction features \cite{salzmann2020trajectron++,lang2019pointpillars,nayakanti2022wayformer,zhou2023query}. The proposed framework only assumes
access to finite-horizon obstacle forecasts. The distribution of the prediction
error is unknown and may change over time due to occlusion, crowd density,
sensing geometry, or changes in obstacle behavior.

\noindent\textbf{Safety requirement.}
Safety with respect to obstacle $j$ is encoded by a margin function
$h_j:\mathcal X\times\mathcal O\to\mathbb R$, where
$h_j(x,o^j)\ge0$ means that robot state $x$ is safe with respect to obstacle
$j$. The multi-obstacle safe set at time $k$ is
\begin{equation}
    \mathcal C_k
    =
    \{x\in\mathcal X: h_j(x,o_k^j)\ge0,\ \forall j\in\mathcal I_k\}.
    \label{eq:safe_set}
\end{equation}
A safe controller should keep the robot inside $\mathcal C_k$ despite uncertain
future obstacle motion. If $o^j_{k+1}$ were known exactly, one-step safety could
be enforced through a discrete-time barrier condition
$h_j(F(x_k,u_k),o^j_{k+1})\ge(1-\gamma)h_j(x_k,o_k^j)$ with
$\gamma\in(0,1]$. In the setting considered here, however, the future obstacle
state is uncertain and its error distribution is not known in advance.

\begin{problem}
Given the robot dynamics \eqref{eq:robot_dynamics}, obstacle observations and
forecasts \eqref{eq:predictor}, and safety set \eqref{eq:safe_set}, design an
online navigation policy for multiple dynamic obstacles with unknown and
non-stationary prediction-error distributions. The policy should maintain task
progress while enforcing risk-aware safety under uncertain obstacle motion and
should avoid committing to local decisions whose predicted future safety risk is
unacceptable. The proposed method addresses this problem without assuming a fixed
parametric uncertainty model for the dynamic environment.
\end{problem}
\section{Adaptive Risk-Certified Event-Triggered Replanning}
\label{sec:method}

\begin{figure}[t]
    \centering
    \includegraphics[width=1.0\linewidth]{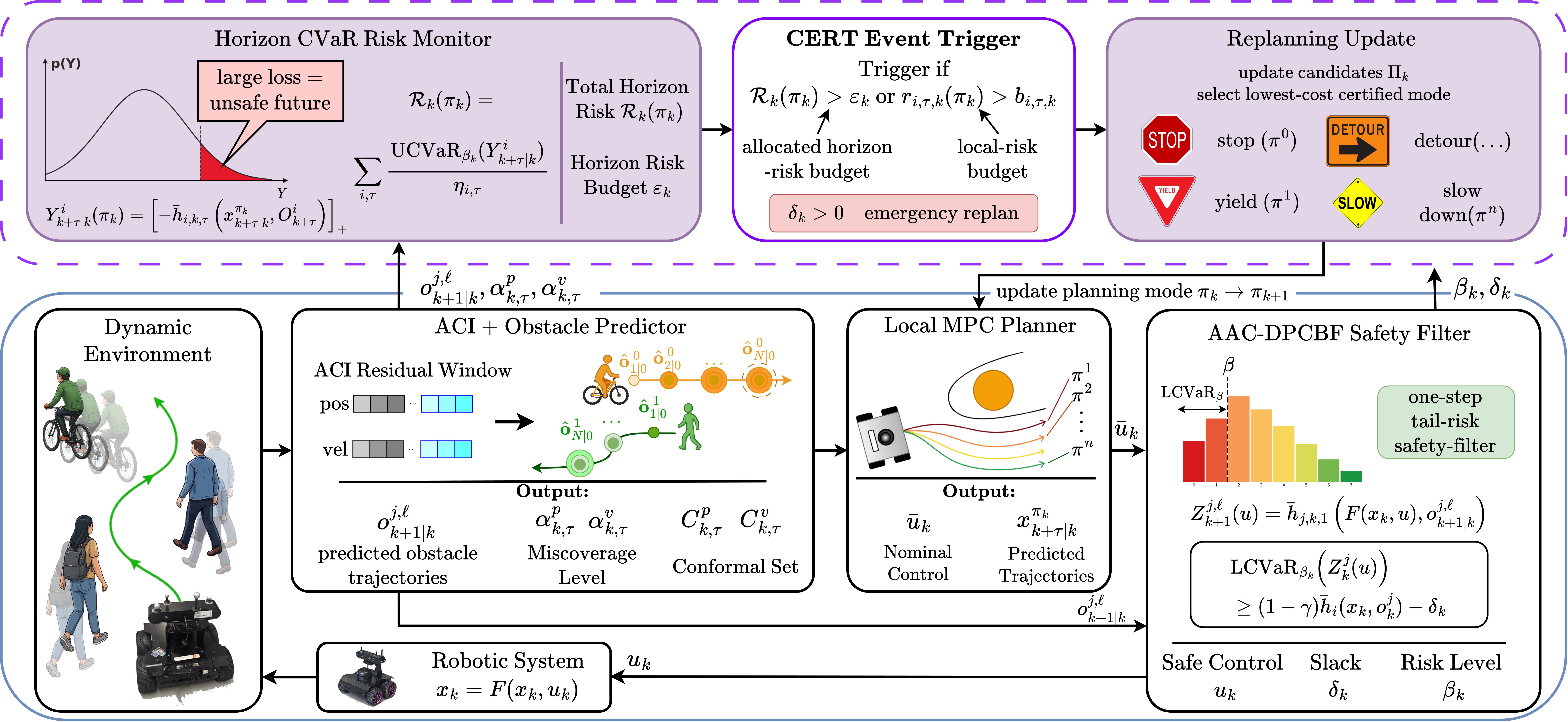}
    \caption{\textbf{CERT-Replan framework.}
The CERT layer monitors the horizon risk of the current local planning mode and
triggers replanning when the mode is no longer risk-certified. The AAC-DPCBF
filter enforces one-step tail-risk-aware safety under calibrated prediction
uncertainty.} %
    \label{fig:framework}
\end{figure}

Figure~\ref{fig:framework} illustrates the proposed framework. At each time step, the robot maintains a local planning mode \(\pi_k\), specifying the reference, speed profile, corridor or homotopy class, and risk allocation used by MPC. We refer to the one-step safety filter as ACI+Adaptive CVaR-DPCBF (AAC-DPCBF), and the full horizon-risk event-triggered replanning framework as CERT-Replan. CERT-Replan evaluates the current mode at two levels: the AAC-DPCBF filter protects the next applied input, while the horizon-risk monitor determines whether the mode should be replaced.

\noindent\textbf{Local mode and rollout.}
Let $\Pi_k$ be the set of candidate modes. Given $\pi\in\Pi_k$, MPC returns a
rollout $x^\pi_{k+\tau|k}$, $\tau=1,\ldots,N$, and nominal first input
$\bar u_k^\pi$. We write $\bar u_k=\bar u_k^{\pi_k}$ for the current mode. Let
$\mathcal F_k$ collect the information available at time $k$, including the robot
state, obstacle tracks, forecasts, residual history, calibrated radii, and
current mode. The random future state of obstacle $j$ conditioned on
$\mathcal F_k$ is denoted by $O^j_{k+\tau}$.

\noindent\textbf{Adaptive conformal calibration.}
For a $\tau$-step forecast made at time $k-\tau$, the realized position and
velocity residuals are
\begin{equation}
    E^p_{j,k,\tau}=\|p^o_{j,k}-\hat p^o_{j,k|k-\tau}\|_2,\qquad
    E^v_{j,k,\tau}=\|v^o_{j,k}-\hat v^o_{j,k|k-\tau}\|_2 .
    \label{eq:prediction_residuals}
\end{equation}
Residuals are computed only for track-matched obstacles. Let
$\mathcal M_{k,\tau}\subseteq\mathcal I_k$ be the set of valid matches, and let
$\mathcal W^q_{k,\tau}=\{E^q_{j,s,\tau}:k-w\le s<k,\ j\in\mathcal M_{s,\tau}\}$
be the sliding residual window for $q\in\{p,v\}$. With adaptive miscoverage
$\alpha^q_{k,\tau}$, the conformal radius is
$C^q_{k,\tau}=\operatorname{Quantile}_{1-\alpha^q_{k,\tau}}(\mathcal W^q_{k,\tau})$.
After observing new residuals, define
$m^q_{k,\tau}=|\mathcal M_{k,\tau}|^{-1}\sum_{j\in\mathcal M_{k,\tau}}
\mathbf 1\{E^q_{j,k,\tau}>C^q_{k,\tau}\}$ and update
$\alpha^q_{k+1,\tau}=\operatorname{proj}_{[\alpha_{\min},\alpha_{\max}]}
(\alpha^q_{k,\tau}+\eta_\alpha(\alpha^\star-m^q_{k,\tau}))$. Thus, the radii
expand under repeated miscoverage and contract when forecasts are over-covered.

\noindent\textbf{Calibrated dynamic-obstacle barrier.}
The nominal margin $h_j(x,o^j)$ is instantiated as a DPCBF depending on
robot--obstacle relative distance and velocity. The calibrated radii define a
tightened margin $\bar h_{j,k,\tau}(x,o^j)=
\mathcal T_j(x,o^j,C^p_{k,\tau},C^v_{k,\tau})$, with 
$\bar h_{j,k,\tau}(x,o^j)\le h_j(x,o^j)$. In our DPCBF implementation,
$C^p_{k,\tau}$ enlarges the effective safety radius and $C^v_{k,\tau}$ makes
relative velocity less favorable. The explicit line-of-sight construction is
given in the appendix. The same tightened margin is used at $\tau=1$ for safety
filtering and over $\tau=1,\ldots,N$ for mode certification.

\noindent\textbf{One-step AAC-DPCBF filter.}
For candidate input $u$, define the random next-step margin
$Z_k^j(u)=\bar h_{j,k,1}(F(x_k,u),O^j_{k+1})$. Since larger margin is safer, the
filter constrains its lower tail:
\begin{equation}
\begin{aligned}
    u_k=\arg\min_{u\in\mathcal U,\delta_k\ge0}\quad
    &\frac12\|u-\bar u_k\|_2^2+c_{\delta}\delta_k^2\\
    \text{s.t.}\quad
    &\operatorname{LCVaR}_{\beta_k}(Z_k^j(u)\mid\mathcal F_k)
    \ge (1-\gamma)\bar h_{j,k,1}(x_k,o_k^j)-\delta_k,\quad
    \forall j\in\mathcal I_k .
\end{aligned}
\label{eq:safety_filter}
\end{equation}
The slack $\delta_k$ is heavily penalized, and $\delta_k>0$ indicates near one-step
filter infeasibility. The tail mass is chosen as the most conservative feasible
value, $\beta_k=\min\{\beta\in[\beta_{\min},\beta_{\max}]:
\eqref{eq:safety_filter}\text{ is feasible with }\delta_k=0\}$. If no such
$\beta$ exists, the least-violating filtered input is used and replanning is
triggered. 
The lower-tail and upper-tail CVaR operators are defined in
Appendix~\ref{app:risk_measures}, and the empirical sample-based
program corresponding to \eqref{eq:safety_filter} is given in
Appendix~\ref{app:empirical_filter}.

\noindent\textbf{Horizon-risk monitor.}
The safety filter certifies the next input but not whether the current mode
should be continued. Let \([a]_+\triangleq \max\{a,0\}\) denote the positive part. For candidate mode $\pi$, obstacle $i$, and horizon step
$\tau$, define the future violation loss
$Y^i_{k+\tau|k}(\pi)=\big[-\bar h_{i,k,\tau}(x^\pi_{k+\tau|k},O^i_{k+\tau})\big]_+$.
For violation scale $\eta_{i,\tau}>0$, define
\begin{equation}
    r_{i,\tau,k}(\pi)=
    \min\left\{1,\,
    \frac{\operatorname{UCVaR}_{\beta_k}
    (Y^i_{k+\tau|k}(\pi)\mid\mathcal F_k)}
    {\eta_{i,\tau}}\right\},
    \qquad
    \mathcal R_k(\pi)=\sum_{\tau=1}^{N}\sum_{i\in\mathcal I_k}
    r_{i,\tau,k}(\pi).
    \label{eq:horizon_risk}
\end{equation}
The same $\beta_k$ couples filtering and monitoring. \begin{proposition}[Conditional horizon-risk certificate]
\label{prop:horizon_risk_certificate}
Fix the information $\mathcal F_k$ and a candidate planning mode $\pi$. For
each obstacle-time pair $(i,\tau)$, let $\eta_{i,\tau}>0$ and define
\[
    \mathcal A^\eta_{i,\tau}(\pi)
    =
    \left\{
    \bar h_{i,k,\tau}
    (x^\pi_{k+\tau|k},O^i_{k+\tau})
    \le
    -\eta_{i,\tau}
    \right\}.
\]
If $r_{i,\tau,k}(\pi)$ and $\mathcal R_k(\pi)$ are defined by
\eqref{eq:horizon_risk}, then
\begin{equation}
    \mathbb P\!\left(
    \bigcup_{\tau=1}^{N}
    \bigcup_{i\in\mathcal I_k}
    \mathcal A^\eta_{i,\tau}(\pi)
    \,\middle|\,
    \mathcal F_k
    \right)
    \le
    \mathcal R_k(\pi).
    \label{eq:horizon_risk_bound}
\end{equation}
\end{proposition}

\noindent
The proof follows from Markov's inequality, the fact that upper-tail CVaR
upper-bounds the expectation of a nonnegative loss, and the union bound~(see
Appendix~\ref{app:horizon_risk_proof} for more details). Proposition~\ref{prop:horizon_risk_certificate}
is a conditional certificate for a fixed candidate mode, not a distribution-free
closed-loop safety guarantee.

\noindent\textbf{CERT trigger and mode update.}
Let $\varepsilon_k$ be the total horizon-risk budget and
$b_{i,\tau,k}$ local budgets with $\sum_{\tau,i}b_{i,\tau,k}\le\varepsilon_k$.
The risk trigger is activated when
\begin{equation}
    \mathcal E_k=
    \mathbf 1\left\{
    \mathcal R_k(\pi_k)>\varepsilon_k
    \ \text{or}\
    \exists i,\tau:\ r_{i,\tau,k}(\pi_k)>b_{i,\tau,k}
    \ \text{or}\
    \delta_k>0
    \right\}.
    \label{eq:cert_trigger}
\end{equation}
\textcolor{black}{To enforce a positive inter-event interval, let $k_m$ denote the most
recent replanning step and define
$n_d=\max\{1,\lceil T_d/\Delta t_e\rceil\}$ for dwell time $T_d>0$
and event-evaluation period $\Delta t_e$. Replanning is governed by
$\mathcal E_k^d=\mathbf 1\{\mathcal E_k=1,\;k-k_m\ge n_d\}$, so replans satisfy
$t_{m+1}-t_m\ge n_d\Delta t_e>0$.}

If $\mathcal E_k^d=0$, the filtered input is applied and $\pi_k$ is retained. If
$\mathcal E_k^d=1$, the planner evaluates candidate modes and selects
\[
    \pi_{k+1}\in\arg\min_{\pi\in\Pi_k}J_k(\pi)
    \quad\text{s.t.}\quad
    \mathcal R_k(\pi)\le\varepsilon_k,\quad
    r_{i,\tau,k}(\pi)\le b_{i,\tau,k}\ \forall i,\tau .
\]
If no candidate is risk-feasible, the least-risk fallback
$\arg\min_{\pi\in\Pi_k}\mathcal R_k(\pi)$ is selected. MPC and the safety filter
are then re-solved under the updated mode before execution.

\begin{algorithm}[t]
\caption{Adaptive risk-certified event-triggered replanning}
\label{alg:cert_replanning}
\begin{algorithmic}[1]
\Require current mode $\pi_k$, residual windows $\{\mathcal W^q_{k,\tau}\}$, budgets $\varepsilon_k,b_{i,\tau,k}$, last replanning step $k_m$, dwell $n_d$
\State Observe $x_k$ and $\{o_k^j\}_{j\in\mathcal I_k}$; forecast obstacle motion
\State Update conformal radii $C^p_{k,\tau},C^v_{k,\tau}$ and tightened margins $\bar h_{j,k,\tau}$
\State Solve MPC under $\pi_k$; solve \eqref{eq:safety_filter} and compute \eqref{eq:horizon_risk}
\State Compute \(\mathcal E_k\) and \(\mathcal E_k^d\)
\If{$\mathcal E_k^d=1$}
    \State Select a lower-risk mode $\pi_{k+1}$ and re-solve MPC/filter
    \State $k_m\gets k$
\EndIf
\State Apply the filtered control $u_k$
\end{algorithmic}
\end{algorithm}

The method treats the local mode as the object being certified: lower-tail risk
checks whether the next control from that mode can be made safe, while upper-tail
horizon risk determines whether the mode should continue to be trusted.
\section{Simulation and Experimental Results}
\label{sec:experiments}

\begin{figure}[t]
    \centering
    \includegraphics[width=1.0\linewidth]{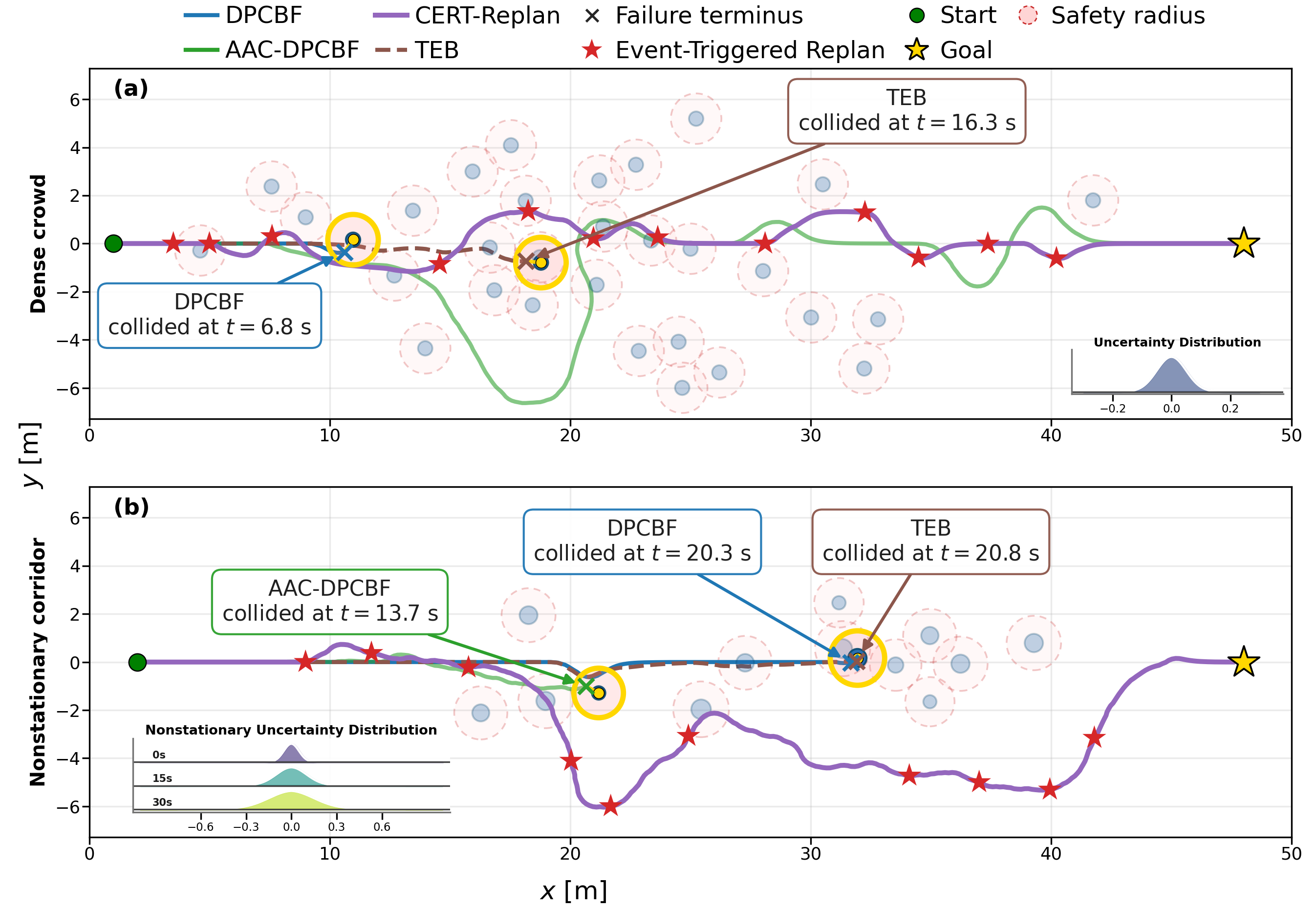}
    \caption{
    Simulation trajectories. (a) Crowded dynamic environment with stationary Gaussian uncertainty. Nominal DPCBF and TEB collide, while AAC-DPCBF and CERT-Replan remain collision-free. Replanning gives modest gains because alternatives have similar risk. (b) Nonstationary corridor with spatially varying uncertainty. The inset shows prediction error shifting from Gaussian to bounded heavy-tailed uncertainty, prompting CERT-Replan to trigger a lower-risk mode change.
    } %
    \label{fig:simulation_distribution}
\end{figure}

\begin{figure}
    \centering
    \includegraphics[width=1.0\linewidth]{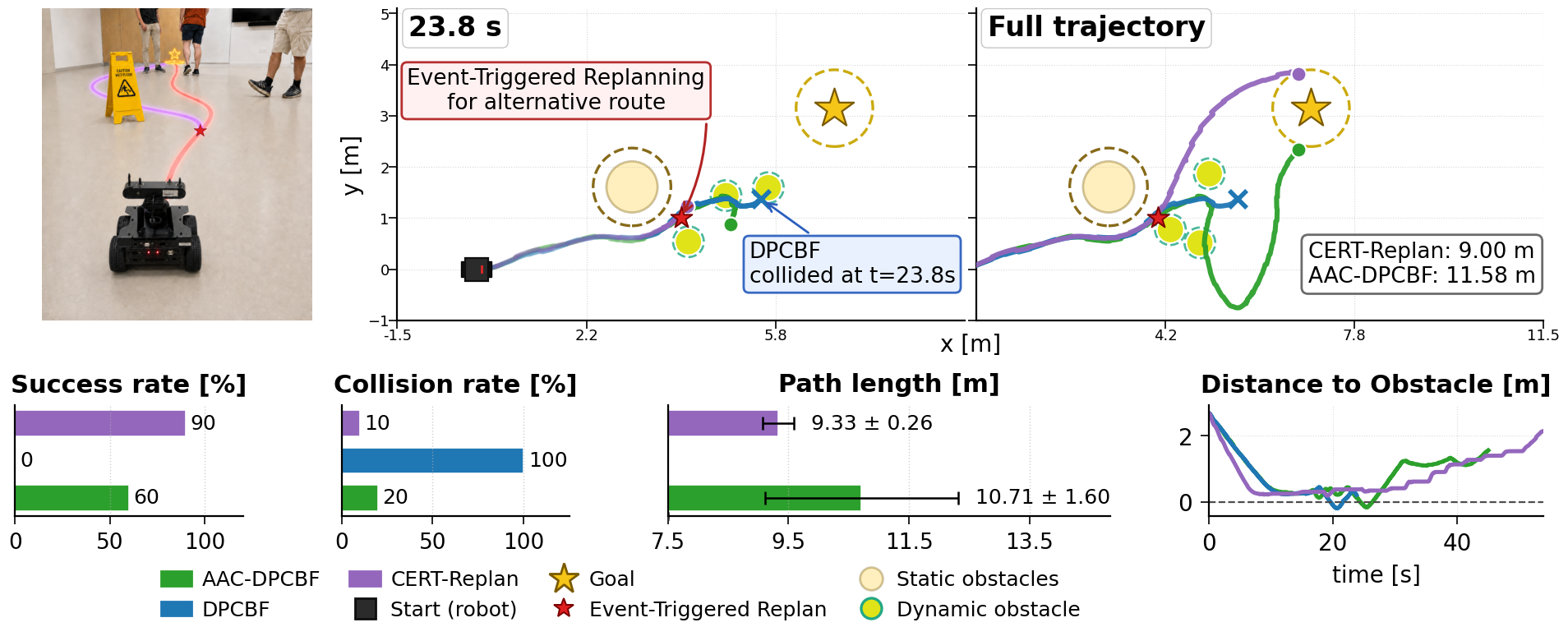}
    \caption{
    Experiment on the Husarion ROSbot 2 Pro over \(10\) trials per method. The robot executes the controller onboard, while dynamic obstacles are virtually injected for repeatability. CERT-Replan achieves the highest success rate and shorter successful-run path length than AAC-DPCBF.
}
    \label{fig:hil_results}
\end{figure}

Figure~\ref{fig:simulation_distribution} illustrates the comparison of CERT-Replan in two simulated scenarios, we additionally evaluate the framework with learned multi-modal prediction, and Figure~\ref{fig:hil_results} shows the hardware experiment. The primary simulations use \(100\) randomized trials per scenario in a \(16\times50\) workspace with moving obstacles whose positions and velocities are subject to uncertainty. The hardware experiment uses \(10\) trials, with dynamic obstacles injected virtually for repeatability. A trial is successful if the robot reaches the goal without violating the safety radius. Otherwise, the trial is counted as a failure due to collision or timeout. Collision indicates a robot-obstacle safety violation, and timeout indicates that the robot exceeds the episode length of \(60\,\mathrm{s}\). When success and collision rates do not sum to \(100\%\), the remaining trials are timeouts. We report path length over successful trials and use \(\|u_k-\bar u_k\|_2\) to measure average safety-filter intervention, where \(\bar u_k\) is the nominal MPC input and \(u_k\) is the filtered input. Additional implementation details, including the vehicle model, obstacle predictor, residual sampling, candidate mode generation, and parameter settings, are provided in Appendix~\ref{app:implementation_details}. Hardware implementation details and experiment-specific parameters are described in Appendix~\ref{app:experiment_details}.

\noindent\textbf{Crowded Dynamic Environment.}
The dense-crowd environment contains \(30\) moving obstacles and tests persistent interaction under prediction noise. Table~\ref{tab:dense-crowd} shows that nominal DPCBF fails frequently, with \(19\%\) success and \(81\%\) collision, while TEB improves success to \(85\%\) but still collides in \(15\%\) of trials. AAC-DPCBF reaches \(96\%\) success and \(4\%\) collision, showing that conformal uncertainty calibration and adaptive tail-risk filtering substantially improve the one-step safety filter. The ablation in Table~\ref{tab:safety-filter-ablation} of Appendix~\ref{app:ablation} further shows that these components are complementary: ACI+DPCBF reduces collision from \(81\%\) to \(20\%\), Adaptive-CVaR-DPCBF improves success to \(76\%\), and their combination gives the strongest safety-filter-only result. CERT-Replan achieves the best overall performance, with \(97\%\) success and \(3\%\) collision, while reducing average safety-filter intervention by $34.2\%$. The gain in success rates over AAC-DPCBF is modest because this scene mainly stresses local obstacle avoidance rather than strategic mode switching; Fig.~\ref{fig:simulation_distribution}(a) shows that nearby alternatives often have similar risk.

\begin{table}[t]
\centering
\caption{Crowded dynamic-environment results over \(100\) randomized trials with \(30\) moving obstacles and unknown obstacle uncertainties. The proposed CERT-Replan achieves the highest success rate and lowest collision rate.}
\label{tab:dense-crowd}
\small
\begin{tabular}{lccccc}
\toprule
Method & Success($\%$) & Collision($\%$) & Path length($m$) & $\|u-\bar{u}_k\|_2$  & Replans \\
\midrule
DPCBF~\cite{park2025beyond}          & 19 & 81 & 47.6$\pm0.7$ & 0.092$\pm0.057$ & - \\
TEB~\cite{rosmann2017integrated}     & 85 & 15 & 47.3$\pm0.4$ & -  & - \\
ICP~\cite{huang2024interaction} & 63 & 37 & 48.2$\pm0.8$ & - & - \\
AAC-DPCBF (Ours)             & 96 & 4 & 51.1$\pm3.2$ & 0.237$\pm0.101$  & -\\
\textbf{CERT-Replan} (Ours)                    & \textbf{97} & \textbf{3} & 51.9$\pm3.2$ & 0.156$\pm0.152$  & 35$\pm3.9$ \\
\bottomrule
\end{tabular}
\end{table}

\noindent\textbf{Non-Stationary Distribution Shift.}
The nonstationary corridor contains \(12\) dynamic obstacles and is designed to isolate the effect of time-varying prediction error on safe navigation. As the robot advances, the noise model changes from a tight Gaussian distribution \((\sigma=0.05)\) to a bounded Gaussian--Laplace mixture with peak tail probability \(0.15\), Laplace scale \(0.25\), and clipping at \(\pm0.75\,\mathrm{m}\). Table~\ref{tab:nonstationary} shows that nominal DPCBF fails under this shift, with \(3\%\) success and \(97\%\) collision. TEB improves success to \(66\%\), but still collides in \(34\%\) of trials. AAC-DPCBF reaches \(79\%\) success and \(12\%\) collision, showing that conformal calibration and adaptive CVaR improve robustness under shifted prediction errors. The ablation in Table~\ref{tab:safety-filter-ablation} of Appendix~\ref{app:ablation} clarifies the role of each component. Adaptive-CVaR-DPCBF without conformal calibration remains close to nominal DPCBF, with only \(4\%\) success under the nonstationary shift, while ACI+DPCBF recovers \(58\%\) success by adapting the uncertainty radii online. Combining both components gives the strongest safety-filter-only baseline, confirming that calibrated uncertainty scaling and adaptive tail-risk filtering are complementary. However, AAC-DPCBF remains a one-step safety filter and can still fail when the current local plan enters regions where uncertainty inflation reduces the feasible control set. 
\textcolor{black}{To isolate the effect of the replanning criterion, we additionally compare CERT-Replan against Slack-Replan and a rate-matched Periodic-Replan. All three replanning methods use the same MPC planner, AAC-DPCBF filter, candidate modes, obstacle realizations, and paired random seeds, and only the trigger differs.}
CERT-Replan achieves \(86\%\) success and \(98\%\) collision-free operation (95\% Wilson CI: \([93.0,99.4]\)), \textcolor{black}{outperforming both Slack-Replan and Periodic-Replan. These gains are significant under paired McNemar tests (\(p\leq0.039\)). Periodic-Replan and CERT-Replan have similar replanning frequencies (\(38.5\) vs. \(39.1\)), indicating that the gain is associated with trigger timing rather than more frequent replanning. CERT-Replan also outperforms the long-horizon conformal-planning baseline ICP, improving success from \(80\%\) to \(86\%\) and reducing collision from \(20\%\) to \(2\%\).}
As shown in Fig.~\ref{fig:simulation_distribution}(b), the risk-certified trigger activates in the transition and heavy-tailed regions, allowing the planner to switch to a lower-risk candidate mode before one-step filtering becomes insufficient. This further demonstrates the benefit of CERT-Replan: when the current local plan accumulates high predicted horizon risk, the framework can switch to a lower-risk candidate mode before one-step filtering becomes insufficient.
\textcolor{black}{Additional sensitivity analyses for the prediction horizon and empirical CVaR sample count are provided in Appendix~\ref{app:sensitivity}.}

\begin{table}[t]
\centering
\caption{Nonstationary distribution-shift results over \(100\) trials with \(12\) moving obstacles. Trigger baselines use paired obstacle realizations and random seeds. The prediction-error distribution changes from Gaussian to bounded heavy-tailed uncertainty.
}
\label{tab:nonstationary}
\small
\begin{tabular}{lccccc}
\toprule
Method & Success($\%$) & Collision($\%$) & Path length($m$) & $\|u-\bar{u}_k\|_2$ & Replans \\
\midrule
DPCBF~\cite{park2025beyond}          & 3 & 97 & 46.0$\pm0.1$ & 0.062$\pm0.045$ & - \\
TEB~\cite{rosmann2017integrated}     & 66 & 34 & 46.2$\pm0.4$ & - & - \\
\textcolor{black}{ICP~\cite{huang2024interaction}} & 80 & 20 & 47.0$\pm0.9$ & - & - \\
AAC-DPCBF (Ours)             & 79 & 12 & 55.7$\pm6.1$ & 0.327$\pm0.121$ & - \\
\textcolor{black}{Slack-Replan} & 83 & 9 & 54.1$\pm7.0$ & $0.234\pm0.066$ & $14.2\pm5.2$ \\
\textcolor{black}{Periodic-Replan} & 81 & 9 & 54.8$\pm5.1$ & $0.298\pm0.078$ & $38.5\pm8.9$ \\
\textbf{CERT-Replan} (Ours)                    & \textbf{86} & \textbf{2} & 53.4$\pm4.1$ & 0.221$\pm0.055$ & 39.1$\pm8.4$ \\
\bottomrule
\end{tabular}
\end{table}

\noindent\textbf{Learned Multi-Modal Prediction.}
\textcolor{black}{
To evaluate CERT-Replan beyond the constant-velocity predictor, we instantiate the same framework with the official Trajectron++ multi-modal predictor~\cite{salzmann2020trajectron++}. All $K=20$ weighted prediction modes are used in both AAC-DPCBF filtering and CERT horizon-risk evaluation. No other component of the safety or replanning framework is changed. Over 100 nonstationary trials, CERT-Replan achieves 96\% collision-free operation and reduces collision incidence by 86.2\% relative to nominal DPCBF and by 91.3\% relative to TEB. These results demonstrate that CERT-Replan operates directly on learned multi-modal prediction samples rather than depending on the constant-velocity model used to isolate the risk-calibration mechanism in the primary experiments.
}

\noindent\textbf{Experiment Results.}
We evaluate CERT-Replan on a Husarion ROSbot 2 Pro with a Jetson Orin Nano. As shown in Fig.~\ref{fig:hil_results}, nominal DPCBF collides in all trials, while AAC-DPCBF improves the success rate to \(60\%\) with a \(20\%\) collision rate but produces the longest successful-run path. CERT-Replan performs best, achieving \(90\%\) success and \(10\%\) collision while reducing successful-run path length relative to AAC-DPCBF. The trajectory snapshots and closest-obstacle distance plot show that CERT-Replan changes its local plan near the high-risk region and maintains positive clearance after the critical interaction, suggesting that the horizon-risk trigger reduces reliance on conservative filter-only corrections by shifting part of the safety response to the planner. Additional representative real-robot snapshots are provided in Fig.~\ref{fig:real_robot_appendix} in Appendix~\ref{app:experiment_details}, and extended videos are included in the supplementary material.

\textcolor{black}{
\noindent\textbf{Onboard Computation.}
We separately profile the complete navigation stack on an onboard Clearpath Jackal computer with a Ryzen 9 8945HX and RTX 4000 Ada. AAC-DPCBF requires \(38.6\pm9.5\,\mathrm{ms}\), CERT monitoring \(38.8\pm1.9\,\mathrm{ms}\), and triggered replanning \(262.0\pm11.3\,\mathrm{ms}\). Across \(1000\) ordinary and \(100\) triggered cycles, no measured cycle misses the \(10\,\mathrm{Hz}\) safety-filter or \(3\,\mathrm{Hz}\) planning deadline; detailed timing statistics are given in Appendix~\ref{app:runtime_profile}.}

\section{Limitations}
\label{sec:limitations}

The finite-horizon risk monitor is a conservative diagnostic based on Markov's inequality, upper-tail CVaR, and a union bound. It certifies the evaluated planning mode, not closed-loop conformal safety. A distribution-free closed-loop guarantee would require additional assumptions on prediction errors, sampling, and replanning. \textcolor{black}{The primary experiments use a constant-velocity predictor to isolate the effect of conformal calibration and risk-certified replanning, while Sec.~\ref{sec:experiments} additionally evaluates the framework with learned multi-modal prediction. In the current simulations, obstacles avoid one another but do not react to the robot. Therefore, mode changes do not alter the residual-generating obstacle policy. Robot-responsive interactions may introduce additional prediction shift after replanning and are not covered by the present conditional certificate.} Future work will evaluate standardized crowd-navigation benchmarks and experiments with physical moving agents.

\section{Conclusion}
\label{sec:conclusion}

We presented CERT-Replan, an event-triggered replanning framework that treats the local planning mode, rather than only the next control input, as the object to be certified under non-stationary obstacle-prediction errors. The central idea is to use calibrated barrier-risk information not only for immediate safety filtering, but also as an early-warning signal for when the current finite-horizon mode should no longer be trusted. By monitoring tail risk along the predicted rollout, CERT-Replan can replace a high-risk mode with a lower-risk alternative before safety filtering becomes overly corrective or infeasible. This shifts dynamic-obstacle navigation from purely reactive control correction toward risk-aware mode management. Simulations and onboard hardware experiments show that this design improves robustness under distribution shift and reduces collisions relative to nominal, safety-filter-only, and \textcolor{black}{simple-trigger} baselines.


\acknowledgments{
This work was supported in part by NASA under Grant 80NSSC23M0219, by an Early Career Faculty grant from NASA’s Space Technology Research Grants Program (Grant 80NSSC26K0696), and by the NVIDIA Academic Grant Program.
}

\clearpage
\bibliography{example}
\appendix

\section{CVaR Operators and Empirical Estimates}
\label{app:risk_measures}

The main text uses lower-tail CVaR for safety margins and upper-tail CVaR for
nonnegative violation losses. For a safety-margin random variable $Z$, larger
values are safer. Its lower-tail CVaR with tail mass $\beta\in(0,1]$ is
\begin{equation}
    \operatorname{LCVaR}_{\beta}(Z\mid\mathcal F_k)
    =
    \sup_{\xi\in\mathbb R}
    \left\{
    \xi
    -
    \frac{1}{\beta}
    \mathbb E
    \big[
    [\xi-Z]_+
    \mid
    \mathcal F_k
    \big]
    \right\}.
    \label{eq:app_lcvar}
\end{equation}
Smaller $\beta$ evaluates a smaller and more adverse lower tail, and is therefore
more conservative.

For a nonnegative loss $Y\ge0$, larger values are less safe. Its upper-tail CVaR
is
\begin{equation}
    \operatorname{UCVaR}_{\beta}(Y\mid\mathcal F_k)
    =
    \inf_{\zeta\in\mathbb R}
    \left\{
    \zeta
    +
    \frac{1}{\beta}
    \mathbb E
    \big[
    [Y-\zeta]_+
    \mid
    \mathcal F_k
    \big]
    \right\}.
    \label{eq:app_ucvar}
\end{equation}
For any nonnegative $Y$,
\begin{equation}
    \mathbb E[Y\mid\mathcal F_k]
    \le
    \operatorname{UCVaR}_{\beta}(Y\mid\mathcal F_k).
    \label{eq:app_ucvar_dominates_expectation}
\end{equation}
This property is used in the proof of Proposition~\ref{prop:horizon_risk_certificate}.

Given samples $\{Z^\ell\}_{\ell=1}^L$ of a safety margin, the empirical
lower-tail CVaR used by the safety filter is
\begin{equation}
    \widehat{\operatorname{LCVaR}}_{\beta}(Z)
    =
    \max_{\xi\in\mathbb R}
    \left\{
    \xi
    -
    \frac{1}{\beta L}
    \sum_{\ell=1}^{L}
    [\xi-Z^\ell]_+
    \right\}.
    \label{eq:app_empirical_lcvar}
\end{equation}
Given samples $\{Y^\ell\}_{\ell=1}^L$ of a nonnegative loss, the empirical
upper-tail CVaR used by the horizon-risk monitor is
\begin{equation}
    \widehat{\operatorname{UCVaR}}_{\beta}(Y)
    =
    \min_{\zeta\in\mathbb R}
    \left\{
    \zeta
    +
    \frac{1}{\beta L}
    \sum_{\ell=1}^{L}
    [Y^\ell-\zeta]_+
    \right\}.
    \label{eq:app_empirical_ucvar}
\end{equation}

\section{DPCBF Geometry and Conformal Tightening}
\label{app:dpcbf_tightening}

This appendix gives the explicit dynamic parabolic CBF used to instantiate the
nominal margin $h_j$ and the tightening operator $\mathcal T_j$ in
Section~\ref{sec:method}.

Let $p_r(x)$ and $v_r(x)$ denote the robot position and velocity. For obstacle
$j$, define the relative position, velocity and signed clearance
\begin{equation}
    r_j(x,o^j)=p^o_j-p_r(x),\qquad
    v_{\mathrm{rel},j}(x,o^j)=v^o_j-v_r(x),\qquad
    d_j(x,o^j)=\|r_j(x,o^j)\|_2-r_{\mathrm{safe}} .
    \label{eq:app_relative_position}
\end{equation}
where $r_{\mathrm{safe}}$ is the required safety radius. The line-of-sight frame
is
\begin{equation}
    e_{\parallel,j}
    =
    \frac{r_j}{\|r_j\|_2+\epsilon},
    \qquad
    e_{\perp,j}
    =
    R_{\pi/2}e_{\parallel,j},
    \qquad
    R_{\pi/2}
    =
    \begin{bmatrix}
        0 & -1\\
        1 & 0
    \end{bmatrix},
    \label{eq:app_los_frame}
\end{equation}
where $\epsilon>0$ avoids division by zero. The relative velocity components are
\begin{equation}
    v_{\parallel,j}
    =
    e_{\parallel,j}^{\top}v_{\mathrm{rel},j},
    \qquad
    v_{\perp,j}
    =
    e_{\perp,j}^{\top}v_{\mathrm{rel},j}.
    \label{eq:app_relative_velocity}
\end{equation}

The nominal DPCBF margin is
\begin{equation}
    h_j(x,o^j)
    =
    v_{\parallel,j}
    +
    \lambda_j(d_j,v_{\mathrm{rel},j})v_{\perp,j}^2
    +
    \mu_j(d_j),
    \label{eq:app_nominal_dpcbf}
\end{equation}
where the curvature and clearance terms are parameterized as
\begin{equation}
    \lambda_j(d_j,v_{\mathrm{rel},j})
    =
    k_\lambda
    \frac{d_j}{\|v_{\mathrm{rel},j}\|_2+\epsilon_v},
    \qquad
    \mu_j(d_j)=k_\mu d_j .
    \label{eq:app_lambda_mu}
\end{equation}
Here \(\epsilon_v>0\) prevents division by zero, \(k_\lambda\) controls the parabolic-boundary curvature, and \(k_\mu\) controls the clearance-dependent margin.

The calibrated radii $C^p_{k,\tau}$ and $C^v_{k,\tau}$ tighten the geometric
quantities entering \eqref{eq:app_nominal_dpcbf}. We use
\begin{equation}
    \bar d_{j,k,\tau}
    =
    \|r_j\|_2-r_{\mathrm{safe}}-C^p_{k,\tau},
    \qquad
    \bar v_{\parallel,j,k,\tau}
    =
    v_{\parallel,j}-C^v_{k,\tau},
    \label{eq:app_tightened_longitudinal}
\end{equation}
and
\begin{equation}
    \bar v_{\perp,j,k,\tau}
    =
    \operatorname{sgn}(v_{\perp,j})
    \max\{|v_{\perp,j}|-C^v_{k,\tau},0\}.
    \label{eq:app_tightened_lateral}
\end{equation}
The tightened DPCBF is
\begin{equation}
    \bar h_{j,k,\tau}(x,o^j)
    =
    \bar v_{\parallel,j,k,\tau}
    +
    \lambda_j(\bar d_{j,k,\tau},v_{\mathrm{rel},j})
    \bar v_{\perp,j,k,\tau}^{2}
    +
    \mu_j(\bar d_{j,k,\tau}).
    \label{eq:app_tightened_dpcbf}
\end{equation}
This construction evaluates the same robot--obstacle state under a smaller
effective clearance and less favorable relative velocity. If a different DPCBF
sign convention is used, the tightening should be applied in the direction that
decreases the safety margin.

\section{Empirical AAC-DPCBF Safety Filter}
\label{app:empirical_filter}

This appendix gives the sample-based program corresponding to
\eqref{eq:safety_filter}. For each obstacle $j$, draw $L$ one-step obstacle
samples $o^{j,\ell}_{k+1|k}$ from the forecast and calibrated residual history.
For a candidate input $u$, define
\begin{equation}
    Z^{j,\ell}_{k+1}(u)
    =
    \bar h_{j,k,1}
    \left(
    F(x_k,u),
    o^{j,\ell}_{k+1|k}
    \right),
    \qquad
    \ell=1,\ldots,L.
    \label{eq:app_sampled_margin}
\end{equation}
The empirical filter solves
\begin{equation}
\begin{aligned}
    \min_{u,\delta_k,\{\xi_j,s_{j\ell}\}}
    \quad&
    \frac12\|u-\bar u_k\|_2^2+c_{\delta}\delta_k^2\\
    \mathrm{s.t.}\quad&
    u\in\mathcal U,\quad \delta_k\ge0,\\
    &s_{j\ell}\ge0,\qquad
    s_{j\ell}\ge \xi_j-Z^{j,\ell}_{k+1}(u),
    \qquad \forall j,\ell,\\
    &\xi_j-\frac{1}{\beta_k L}\sum_{\ell=1}^{L}s_{j\ell}
    \ge
    (1-\gamma)\bar h_{j,k,1}(x_k,o_k^j)-\delta_k,
    \qquad \forall j\in\mathcal I_k .
\end{aligned}
\label{eq:app_empirical_filter}
\end{equation}
The variables $\xi_j$ and $s_{j\ell}$ implement the empirical lower-tail CVaR constraint. Specifically, the lower-tail empirical CVaR contains the hinge term $\big[\xi_j-Z^{j,\ell}_{k+1}(u)\big] _+$. Introducing $s_{j\ell}$ with $s_{j\ell}\ge \xi_j-Z^{j,\ell}_{k+1}(u)$ and $s_{j\ell}\ge0$ gives the standard reformulation of this hinge term, which is commonly used to express CVaR constraints in optimization programs~\cite{sarykalin2008value}. If \eqref{eq:app_empirical_filter} is nonlinear in $u$, we solve it by sequential convexification around the nominal MPC command or rollout.

The adaptive tail mass $\beta_k$ is selected on a finite grid in
$[\beta_{\min},\beta_{\max}]$. The filter starts from the most conservative
candidate and increases $\beta$ only as needed to recover feasibility with
$\delta_k=0$. If no grid value is feasible without slack, the least-violating
solution is returned and the CERT trigger is activated.

\section{Proof of the Conditional Horizon-Risk Certificate}
\label{app:horizon_risk_proof}

We prove Proposition~\ref{prop:horizon_risk_certificate}. Fix $\mathcal F_k$ and
a candidate planning mode $\pi$. For obstacle $i$ and horizon step $\tau$, recall
the loss
\[
    Y^i_{k+\tau|k}(\pi)
    =
    \left[
    -
    \bar h_{i,k,\tau}
    \left(
    x^\pi_{k+\tau|k},
    O^i_{k+\tau}
    \right)
    \right]_+
\]
and the margin-violation event
\[
    \mathcal A^\eta_{i,\tau}(\pi)
    =
    \left\{
    \bar h_{i,k,\tau}
    \left(
    x^\pi_{k+\tau|k},
    O^i_{k+\tau}
    \right)
    \le
    -\eta_{i,\tau}
    \right\}.
\]
If $\mathcal A^\eta_{i,\tau}(\pi)$ occurs, then
$Y^i_{k+\tau|k}(\pi)\ge\eta_{i,\tau}$. Markov's inequality gives
\begin{equation}
    \mathbb P
    \left(
    \mathcal A^\eta_{i,\tau}(\pi)
    \mid
    \mathcal F_k
    \right)
    \le
    \frac{
    \mathbb E
    [
    Y^i_{k+\tau|k}(\pi)
    \mid
    \mathcal F_k
    ]
    }
    {\eta_{i,\tau}}.
    \label{eq:app_markov}
\end{equation}
Since $Y^i_{k+\tau|k}(\pi)\ge0$, upper-tail CVaR upper-bounds its conditional
expectation. Hence,
\begin{equation}
    \mathbb P
    \left(
    \mathcal A^\eta_{i,\tau}(\pi)
    \mid
    \mathcal F_k
    \right)
    \le
    \frac{
    \operatorname{UCVaR}_{\beta_k}
    \left(
    Y^i_{k+\tau|k}(\pi)
    \mid
    \mathcal F_k
    \right)
    }
    {\eta_{i,\tau}}
    \le
    r_{i,\tau,k}(\pi),
    \label{eq:app_single_bound}
\end{equation}
where the final inequality follows from the definition of $r_{i,\tau,k}(\pi)$,
with truncation at one. Applying the union bound over all obstacles and horizon
steps gives
\begin{equation}
    \mathbb P
    \left(
    \bigcup_{\tau=1}^{N}
    \bigcup_{i\in\mathcal I_k}
    \mathcal A^\eta_{i,\tau}(\pi)
    \mid
    \mathcal F_k
    \right)
    \le
    \sum_{\tau=1}^{N}
    \sum_{i\in\mathcal I_k}
    r_{i,\tau,k}(\pi)
    =
    \mathcal R_k(\pi).
    \label{eq:app_union_bound}
\end{equation}
This proves the proposition.

\section{Safety-Filter Ablation Studies}
\label{app:ablation}

We perform an ablation study to isolate the contribution of each safety-filter component before adding risk-certified replanning. All methods use the same MPC planner, robot dynamics, DPCBF parameters, and obstacle scenarios. They differ only in how the DPCBF constraint handles prediction uncertainty and tail risk. The nominal DPCBF baseline uses the predicted obstacle state without conformal tightening or CVaR risk filtering. ACI+DPCBF adds adaptive conformal inference to the nominal DPCBF by tightening the barrier with calibrated position and velocity radii, but does not use CVaR. Adaptive-CVaR-DPCBF uses adaptive CVaR tail-risk filtering without conformal prediction or conformal tightening. AAC-DPCBF combines both components, using adaptive conformal calibration together with adaptive CVaR-DPCBF filtering. This ablation isolates the effect of conformal uncertainty calibration, adaptive CVaR tail-risk filtering, and their interaction under otherwise identical planning and control settings.

\begin{table}[t]
\centering
\caption{
Safety-filter ablation without risk-certified replanning. 
All methods use the same robot dynamics, MPC horizon, DPCBF parameters, and obstacle scenarios as the main experiments.
}
\label{tab:safety-filter-ablation}
\small
\begin{tabular}{llcccc}
\toprule
Environment & Method & Success(\%) & Collision(\%) & Path length(m) & $\|u-\bar{u}_k\|_2$ \\
\midrule
\multirow{4}{*}{Dense crowd}
& DPCBF~\cite{park2025beyond}          & 19 & 81 & 47.6$\pm0.7$ & 0.092$\pm0.057$ \\
& ACI+DPCBF                            & 71 & 20 & 49.3$\pm2.2$ & 0.474$\pm0.327$ \\
& Adaptive-CVaR-DPCBF                  & 76 & 24 & 47.7$\pm0.9$ & 0.109$\pm0.051$ \\
& AAC-DPCBF                            & 96 & 4  & 51.1$\pm3.2$ & 0.237$\pm0.101$ \\
\midrule
\multirow{4}{*}{Nonstationary}
& DPCBF~\cite{park2025beyond}          & 3  & 97 & 46.0$\pm0.1$ & 0.062$\pm0.045$ \\
& ACI+DPCBF                            & 58 & 37 & 49.2$\pm2.3$ & 0.613$\pm0.335$ \\
& Adaptive-CVaR-DPCBF                  & 4  & 96 & 45.8$\pm0.1$ & 0.062$\pm0.038$ \\
& AAC-DPCBF                            & 79 & 12 & 55.7$\pm6.1$ & 0.327$\pm0.121$ \\
\bottomrule
\end{tabular}
\end{table}

Table~\ref{tab:safety-filter-ablation} shows that both ACI and adaptive CVaR improve over nominal DPCBF, but their roles differ. In the dense-crowd setting, ACI+DPCBF strongly reduces collision by inflating the barrier under prediction uncertainty, while Adaptive-CVaR-DPCBF improves success with smaller average control intervention by focusing the filter on lower-tail barrier risk. Combining both gives the strongest safety-filter-only result, improving success to \(96\%\) and reducing collision to \(4\%\).

The nonstationary setting highlights the importance of calibration. Adaptive-CVaR-DPCBF without ACI performs similarly to nominal DPCBF, indicating that risk adaptation alone cannot compensate for a shifted or miscalibrated prediction-error distribution. ACI+DPCBF recovers much of the lost performance by adapting the uncertainty radii online, and AAC-DPCBF further improves success to \(79\%\) while reducing collision to \(12\%\). These results support the design of the proposed safety filter: conformal calibration provides reliable uncertainty scaling, while adaptive CVaR shapes the controller response to tail-risk events.

\section{Implementation Details}
\label{app:implementation_details}

\noindent\textbf{Vehicle Dynamics.}
The simulation uses a discrete-time kinematic bicycle model. The robot state is \(x_k=[p_{x,k},p_{y,k},\theta_k,v_k]^\top\), where \(p_{x,k},p_{y,k}\) denote position, \(\theta_k\) is heading, and \(v_k\) is forward speed. The control input is \(u_k=[a_k,\rho_k]^\top\), where \(a_k\) is acceleration and \(\rho_k\) is the steering angle. With sampling time \(\Delta t\) and wheelbase \(L_{\mathrm{wb}}\), the dynamics are discretized using forward Euler integration:
\begin{equation} 
    \begin{aligned} 
        p_{x,k+1} &= p_{x,k} + \Delta t\, v_k\cos\theta_k,\\ 
        p_{y,k+1} &= p_{y,k} + \Delta t\, v_k\sin\theta_k,\\ 
        \theta_{k+1} &= \theta_k + \Delta t\,\frac{v_k}{L_{\mathrm{wb}}}\tan\rho_k,\\ 
        v_{k+1} &= v_k + \Delta t\, a_k . 
    \end{aligned} 
    \label{eq:app_bicycle_dynamics} 
\end{equation}
The inputs are constrained as \(a_k\in[-2,+2]\) and \(\rho_k\in[-0.6,0.6]\), and the speed is clipped to \(v_k\in[0.1,1.5]\). This model is used for all simulation baselines so that differences in performance are due to the safety and replanning layers rather than the robot dynamics.

\noindent\textbf{Obstacle predictor.}
The framework is predictor-agnostic. In the primary experiments, unless otherwise
specified, we use a constant-velocity predictor:
\begin{equation}
    \hat p^o_{j,k+\tau|k}
    =
    p^o_{j,k}
    +
    \tau\Delta t\,v^o_{j,k},
    \qquad
    \hat v^o_{j,k+\tau|k}
    =
    v^o_{j,k}.
    \label{eq:app_constant_velocity}
\end{equation}
\textcolor{black}{For the learned multi-modal evaluation in Sec.~\ref{sec:experiments}, we instantiate the same predictor interface with Trajectron++~\cite{salzmann2020trajectron++} and retain all \(K=20\) weighted prediction modes for AAC-DPCBF filtering and CERT horizon-risk evaluation. The safety and replanning formulations are otherwise unchanged.}

\noindent\textbf{Residual sampling.}
For empirical CVaR evaluation, obstacle samples are generated by perturbing the
nominal forecast with residual samples from the corresponding horizon-indexed
windows. For horizon step $\tau$, position and velocity perturbations are sampled
from $\mathcal W^p_{k,\tau}$ and $\mathcal W^v_{k,\tau}$, respectively.


\noindent\textbf{Candidate modes.}
The candidate mode set \(\Pi_k\) contains local alternatives to the current MPC mode. In our implementation, candidates are generated by modifying the nominal reference through speed scaling, lateral-offset detours, and a closest-obstacle-behind detour. The speed-scaled modes use \(v^{\mathrm{ref}}\in\{1.0,0.75,0.5,0.25,0\}v^{\mathrm{ref}}_{\mathrm{nom}}\), where the zero-speed mode corresponds to yielding. The lateral-offset modes shift the local reference by \(\pm 1.0\,\mathrm{m}\) and \(\pm 1.5\,\mathrm{m}\) relative to the nominal path when a feasible corridor exists. The closest-obstacle-behind mode constructs a detour that passes behind the nearest high-risk obstacle, where ``behind'' is defined with respect to the obstacle's predicted velocity direction. Each candidate is evaluated by solving MPC and computing the horizon risk \(\mathcal R_k(\pi)\). If at least one candidate satisfies the CERT constraints, the lowest-cost certified candidate is selected. If no candidate is certified, the fallback mode is the candidate with minimum \(\mathcal R_k(\pi)\).

\noindent\textbf{Simulation Parameters.}
Table~\ref{tab:sim_params} lists the main parameters used in the simulation study. The same values are used across all methods unless otherwise specified. The ACI residual window is initialized with \(n_{\mathrm{init}}\) synthetic residual samples drawn from the initial prediction-error distribution, where the remaining entries are initialized as zero residuals.

\begin{table}[t]
\centering
\caption{Simulation parameters used across methods.}
\label{tab:sim_params}
\small
\begin{tabular}{lll}
\toprule
Parameter & Value & Description \\
\midrule
\(\Delta t\) & \(0.1\,\mathrm{s}\) & Simulation time step \\
\(L_{\mathrm{wb}}\) & \(0.5\,\mathrm{m}\) & Bicycle-model wheelbase \\
\(\gamma\) & \(0.4\) & Discrete-time DPCBF rate \\
\(r_{\mathrm{safe}}\) & \(0.75\,\mathrm{m}\) & Base robot-obstacle safety radius \\
\(k_\lambda\) & \(1.0\) & DPCBF curvature term \\
\(k_\mu\) & \(1.0\) & DPCBF clearance-margin term \\
\(N\) & \(6\) & MPC and prediction horizon \\
\(\alpha^\star\) & \(0.1\) & Target ACI miscoverage \\
\(\eta_\alpha\) & \(0.01\) & ACI adaptation gain \\
\(w\) & \(256\) & Residual window length \\
\(n_{\mathrm{init}}\) & \(64\) & Warm-start residual samples \\
\(N_s\) & \(16\) & Empirical CVaR samples \\
\([\beta_{\min},\beta_{\max}]\) & \([0.05,0.30]\) & Adaptive CVaR range \\
\(N_\beta\) & \(6\) & CVaR grid size \\
\(N_{\mathrm{obs}}\) & \(10\) & Maximum constrained obstacles \\
\bottomrule
\end{tabular}
\end{table}

\section{Experiment Details}
\label{app:experiment_details}

\begin{figure}
    \centering
    \includegraphics[width=1.0\linewidth]{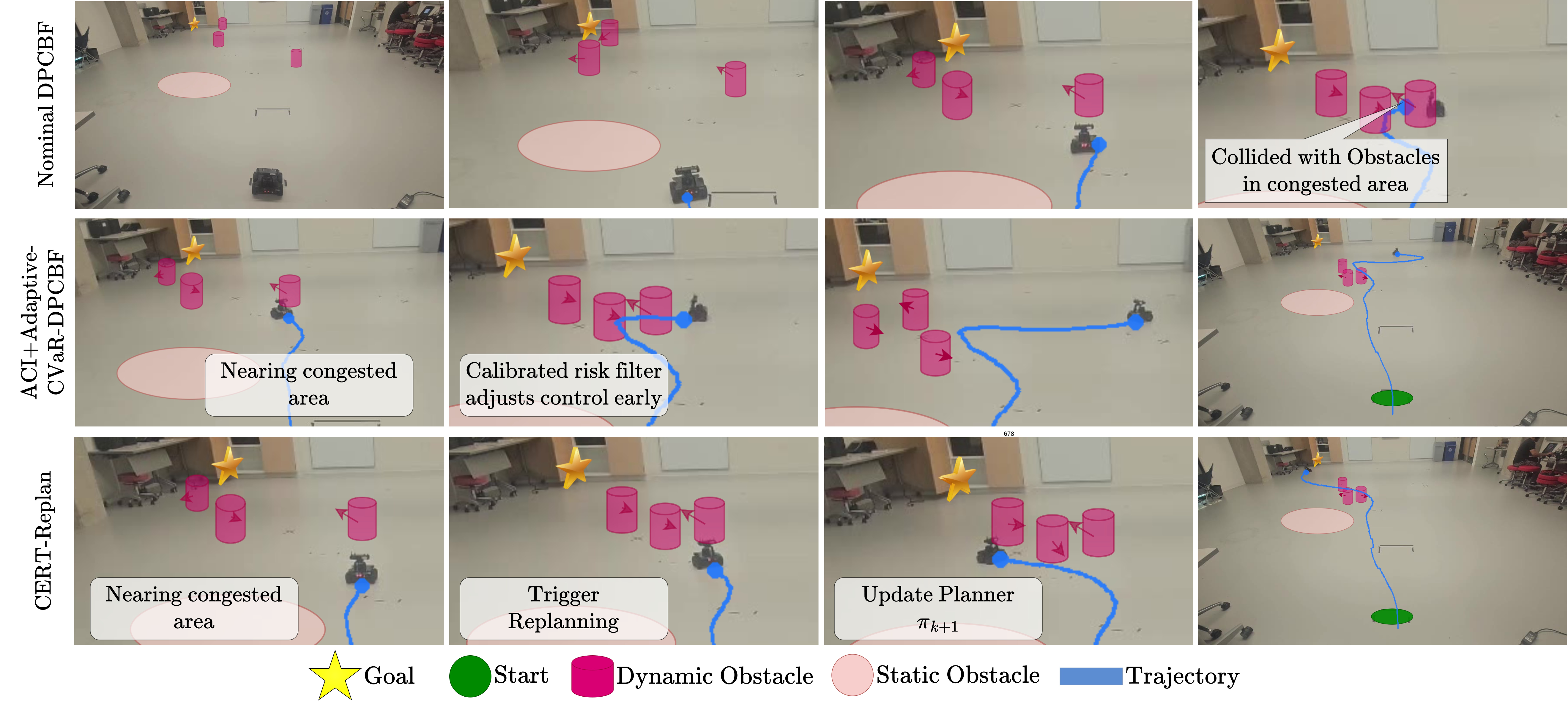}
    \caption{Real-robot experiment on the Husarion ROSbot 2 Pro. 
    Rows compare nominal DPCBF, ACI+Adaptive-CVaR-DPCBF, and CERT-Replan across successive snapshots. 
    The robot physically executes the onboard controller, while dynamic obstacles and prediction uncertainty are injected virtually. 
    Nominal DPCBF collides with the dynamic obstacles, whereas CERT-Replan changes its local plan and reaches the goal.
    }
    \label{fig:real_robot_appendix}
\end{figure}

Figure~\ref{fig:real_robot_appendix} shows the real-robot evaluation setup and representative snapshots for each method. The robot physically executes the onboard controller on the Husarion ROSbot 2 Pro, while the dynamic obstacles and prediction uncertainty are injected virtually for repeatability. The figure illustrates the qualitative behavior observed in the trials: nominal DPCBF collides with the dynamic obstacles, ACI+Adaptive-CVaR-DPCBF avoids collision through safety-filter corrections, and CERT-Replan changes the local plan to pass the high-risk congested region and reach the goal.

\noindent\textbf{Control Command.}
The real-robot experiment is performed on a Husarion ROSbot 2 Pro equipped with a Jetson Orin Nano. The planning and safety stack uses the same kinematic bicycle model as in simulation, while the ROSbot executes differential-drive velocity commands. Therefore, the filtered bicycle-model command \(u_k=[a_k,\rho_k]^\top\) is converted to a ROS~2 \texttt{/cmd\_vel} command through a velocity profiler. The nominal MPC runs at \(3\,\mathrm{Hz}\), and the AAC-DPCBF safety filter runs at \(10\,\mathrm{Hz}\). At each safety-filter update, the filtered acceleration command is integrated to obtain a desired forward speed,
\begin{equation}
    v^{\mathrm{des}}_{k+1} = \operatorname{clip} \left( v_k+\Delta t_{\mathrm{sf}} a_k, v_{\min}, v_{\max} \right),
\end{equation}
where \(\Delta t_{\mathrm{sf}}=0.1\,\mathrm{s}\). A low-level velocity profiler then runs at \(100\,\mathrm{Hz}\) and tracks \(v^{\mathrm{des}}_{k+1}\) subject to acceleration limits, producing a smooth commanded speed \(v_{\mathrm{cmd}}\).

The steering angle is converted to differential-drive angular velocity using the kinematic bicycle relation
\begin{equation}
    \omega_{\mathrm{cmd}} = \frac{v_{\mathrm{cmd}}}{L_{\mathrm{wb}}}\tan(\rho_k).
    \label{eq:ackermann_to_unicycle}
\end{equation}
Both \(v_{\mathrm{cmd}}\) and \(\omega_{\mathrm{cmd}}\) are clipped to the ROSbot actuation limits before execution. This conversion allows the same high-level MPC and safety-filter stack to be executed on the differential-drive platform while preserving the commanded curvature of the bicycle-model trajectory.

\noindent\textbf{Mapping and Localization.}
The testing area is mapped with SLAM before the experiment, and AMCL is used for localization during each trial. Static obstacles are known from the map and used to generate the initial reference path. Dynamic obstacles are provided to the planner only when their distance to the robot is below \(3\,\mathrm{m}\), so the robot operates with local obstacle information. Localization error contributes to the effective robot-obstacle relative-state error and is therefore reflected in the calibrated residuals used by the safety layer.

\noindent\textbf{Experiment Parameters.}
The onboard experiment uses the same parameters as the simulation unless otherwise specified. To accommodate the lower update rate and onboard computation limits, we use \(N_{\mathrm{obs}}=4\), \(N_{\beta}=3\), \(\Delta t=0.25\,\mathrm{s}\), \(r_{\mathrm{safe}}=0.23\,\mathrm{m}\), and \(N=5\).



\noindent\textbf{Onboard Runtime Profiling}
\label{app:runtime_profile}
We profile the complete CERT-Replan stack on an onboard Clearpath Jackal computer equipped with a Ryzen 9 8945HX and RTX 4000 Ada. AAC-DPCBF requires \(38.6\pm9.5\,\mathrm{ms}\) per update (p95: \(65.6\,\mathrm{ms}\)), CERT horizon-risk monitoring requires \(38.8\pm1.9\,\mathrm{ms}\) (p95: \(40.6\,\mathrm{ms}\)), and a triggered replanning cycle requires \(262.0\pm11.3\,\mathrm{ms}\) (p95: \(278.8\,\mathrm{ms}\)). Across \(1000\) ordinary cycles and \(100\) triggered replanning cycles, no measured cycle misses the \(10\,\mathrm{Hz}\) safety-filter or \(3\,\mathrm{Hz}\) planning deadline.

\section{Sensitivity Analysis}
\label{app:sensitivity}
\noindent\textbf{Prediction horizon.}
We evaluate CERT-Replan over \(10\) trials per setting with \(N_s=32\) and \(N\in\{4,6,8,10,12\}\). Success rates are \(30\%\), \(40\%\), \(50\%\), \(80\%\), and \(60\%\), respectively, while collision-free rates are \(80\%\), \(60\%\), \(70\%\), \(80\%\), and \(100\%\). At \(N=12\), all trials remain collision-free, but \(4/10\) time out, illustrating that longer horizons can increase conservatism as predictive risk accumulates.
\[
\begin{array}{c|ccccc}
N & 4 & 6 & 8 & 10 & 12\\
\hline
\text{Success (\%)} & 30 & 40 & 50 & 80 & 60\\
\text{Collision-free (\%)} & 80 & 60 & 70 & 80 & 100
\end{array}
\]
\noindent\textbf{CVaR sample count.}
With \(N=12\), we sweep \(N_s\in\{16,32,64,128,512\}\). Mean horizon risk is \(34.88\), \(37.84\), \(37.22\), \(37.35\), and \(39.38\), respectively, indicating that the risk estimate is comparatively stable from \(N_s=32\) onward for the evaluated prediction-error distribution. Mean replans are \(46.8\), \(42.5\), \(40.0\), and \(40.4\) for \(N_s=32,64,128,512\), respectively.
\[
\begin{array}{c|ccccc}
N_s & 16 & 32 & 64 & 128 & 512\\
\hline
\text{Mean horizon risk} & 34.88 & 37.84 & 37.22 & 37.35 & 39.38\\
\text{Mean replans} & {-} & 46.8 & 42.5 & 40.0 & 40.4
\end{array}
\]The required sample count remains distribution-dependent. These sweeps are sensitivity analyses; the primary benchmark configuration remains the one reported in Table~\ref{tab:sim_params}.

\end{document}